# Dynamic Capitalization and Visualization Strategy in Collaborative Knowledge Management System for EI Process

Bolanle F. Oladejo, Victor T. Odumuyiwa and Amos A. David

*Abstract*—Knowledge is attributed to human whose problem-solving behavior is subjective and complex. In today's knowledge economy, the need to manage knowledge produced by a community of actors cannot be overemphasized. This is due to the fact that actors possess some level of tacit knowledge which is generally difficult to articulate. Problem-solving requires searching and sharing of knowledge among a group of actors in a particular context. Knowledge expressed within the context of a problem resolution must be capitalized for future reuse. In this paper, an approach that permits dynamic capitalization of relevant and reliable actors' knowledge in solving decision problem following Economic Intelligence process is proposed. Knowledge annotation method and temporal attributes are used for handling the complexity in the communication among actors and in contextualizing expressed knowledge. A prototype is built to demonstrate the functionalities of a collaborative Knowledge Management system based on this approach. It is tested with sample cases and the result showed that dynamic capitalization leads to knowledge validation hence increasing reliability of captured knowledge for reuse. The system can be adapted to various domains.

## I. INTRODUCTION

IN today's knowledge economy, the need to manage knowledge produced by a community of actors cannot be overemphasized. This is due to the fact that actors possess some level of tacit knowledge which is generally difficult to articulate. In fact it has been said that users possess some knowledge which they themselves may not be aware of until they are faced with real problem that will steer up such knowledge in them [1]. Such knowledge expressed within the context of a problem resolution must be capitalized for future reuse. The problem solving method of a group of actors and the resulting solution to their problem are forms of knowledge that can be capitalized. Due to the inherent advantages of capitalizing actors' knowledge, it is inevitable to initiate a Knowledge Management (KM) approach which incites externalization of such knowledge and dynamically track their communication. This work thus aims at initiating a collaborative KM approach which applies capitalization technique and method that facilitate dynamic, reliable, and innovative creation of knowledge.

The context of this work focuses on decision problems resolution in organizations. Decision making (DM) is an inevitable task in virtually every organization. Since every organization is goal-oriented, a strategic approach to DM will be of paramount importance. We are interested in Economic Intelligence (EI) process which facilitates effective DM through collection, treatment and use of relevant information by actors in resolving decision problem. In DM, we identify a decision problem and then take decision from the various alternatives available. From the starting point of identifying a decision problem to the point of taking a decision, there are various processes involved. All these are taken care of in EI [2].

The next section considers a brief theoretical background and summary of related works. The third and fourth sections discuss the recursive capitalization approach and its implementation in our KM system. The paper is concluded in the fifth section.

## II. REVIEW OF RELATED CONCEPTS AND APPLICATIONS

### A. Overview of Economic Intelligence (EI) Process

EI focuses on the use of information to solve *Decision making (DM)* problems. It concerns the set of concepts, methods and tools which unify all the coordinated actions of research, acquisition, treatment, storage and diffusion of information relevant to individual or clustered enterprises and organizations in the framework of a strategy [3]. The goal of EI is to reduce uncertainty in decision making. EI process is made up of the following phases:
  ✓ Decision problem identification and definition
  ✓ Information gathering
  ✓ Adequacy of information
  ✓ Protection of information heritage
  ✓ Use of information.

B. F. Oladejo is with SITE research team, LORIA, Campus Scientifique, Nancy University, Vandoeuvre les Nancy Cedex Nancy, France (phone: 383-592-087; e-mail: oladejof@loria.fr).
V. T. Odumuyiwa is with SITE research team, LORIA, Campus Scientifique, Nancy University, France (e-mail: victor.odumuyiwa@ loria.fr).
A. A. David is the head of SITE research team, LORIA, Campus Scientifique, Nancy, France (e-mail: amos.david@loria.fr).

Before elaborating these phases, it is necessary to present the main actors that are involved in execution of the process. They are as listed below.

Decision maker: this actor is the one capable of identifying and establishing the problem to be solved in terms of stake, of risk or threat on the enterprise [4]. In other words, he knows the needs of the organization, the stakes, the eventual risks and the threats the organization can be subjected to [5].

Information watcher: This actor is responsible for locating, supervising, validating, and emphasizing the strategic information needed for solving decision problem. He works hand in hand with the decision maker right from the initial stage of making a decision problem explicit. He translates this problem into information search problem so as to begin the collection of relevant information for solving the problem.

The above-mentioned actors are involved in the various phases of EI process. Subsequently, summary of the phases are presented.

The starting point in EI process is the identification of decision problem (DP). This aims at defining a DP and its associated contexts in order to assess the scope of the DP. It is handled by a set of decision makers or policy makers who are charged with the responsibility of clarifying the concepts or terms in a DP. They also observe the internal and external environments of the concerned organization in order to determine the parameters associated with identified DP. The observation leads to formulation of assumptions that address the challenge of risk or likely threat (what is likely to be won or lost) associated to the DP if not appropriately handled. This phase includes a validation activity whereby the actors verify the concepts, clarify the stated assumptions and verify any related laws or theorems. The assumptions are checked in order to confirm ideas with respect to the parameters of the organization. This phase facilitates better understanding of the information needs of decision makers by information watchers. The phase is succeeded by the process of information search and gathering.

The watcher is concerned with the process of information search based on his understanding of the decision maker's problem which has been defined and validated. In order for him to target the central aim of EI, that is, provision of relevant information to users' needs, he needs to identify and validate appropriate information search requirements that actually address the DP and its contexts. DP is translated into precise and adequate Information Search Problem (ISP) which will help to identify relevant information sources and to determine the assessment criteria for adequacy of the required information.

Consequently, there is need to validate the adequacy of ISP with respect to DP before proceeding to information gathering. This requires strategic and conscious assessment of concepts and specification of the ISP by watchers [4] and decision makers. It is after a concession is reached before the ISP could be accepted. Subsequently, the relevant sources of information are identified and validated. The collection of relevant information is based on selection from information world according to the specification of ISP. This information cannot be used directly by decision makers. Thus, the actors collaborate to analyze the validated information and to identify indicators that can be used to guide the required decision. The indicators are interpreted by the actor before decision is taken [6]. It is important to note that all the expressed and collected information in EI process should be properly protected from unwise divulgation as well as from spies and competitors [2]. This constitutes the phase of protection of information heritage. All the information and communication that ensue from the collaborating actors in the course of solving DP constitute knowledge.

### B. The Concept - Knowledge

Knowledge is regarded as a capital which has an economic value [7]. Knowledge is also considered as a strategic resource for gaining competitive advantage and for innovation in organizations. From related literature, it has been distinguished from data and information [8], [9]. Hitherto, there is no universal definition of knowledge. In the context of this work, knowledge is defined as facts with its attributed meaning, where meaning is a function of an observation, learning, experience, and understanding of a reality in a particular situation or context at a specific period of time by an individual [9]. Thus, knowledge is inseparable from individuals and it is reflected in the role designated to them.

There are two main types of knowledge, explicit (objective) and tacit (subjective) knowledge [10]. The various kinds of knowledge are illustrated in figure 1. They are referred to as knowledge resources in this work. Tacit knowledge refers to 'know-how' of an individual while explicit knowledge is the articulated knowledge in form of documents, operation manual, video, etc. The latter could be readily transmitted across individuals formally and systematically. However, tacit knowledge requires a transformation process to convert it to explicit knowledge before it can be capitalized. According to Nonaka, there are four knowledge conversion processes for transforming tacit to explicit knowledge and vice-visa [10]. These include:

- Socialization: a process whereby tacit knowledge is converted to tacit knowledge. This can take place in a discussion forum, team meeting etc. In EI context, socialization occurs in the course of communication among collaborating actors.
- Externalization: enables the transformation of tacit knowledge to explicit knowledge. In a situation where an individual expresses his knowledge in form of a document or experience report, he is actually converting his tacit knowledge to explicit knowledge. In EI context, externalization occurs during decision problem elicitation.
- Internalization: a process whereby explicit knowledge is converted to tacit knowledge. A reading process is an example of internalization whereby a reader acquires tacit knowledge from an explicit knowledge artifact. In EI context, internalization occurs during the analysis of collected relevant information.

▪ Combination: a process whereby explicit knowledge is transformed to explicit knowledge. For instance, a document can be indexed to facilitate its retrieval. The words used to index such document can be considered as explicit knowledge on the document. Combination occurs in EI context during the generation of indicators from the collected relevant information.

Table 1 summarizes the four conversion processes and as well states instances in EI process through which the conversion takes place. Generated knowledge through these processes should be managed with the aim of facilitating its reuse [11]. Since we aim at capitalization of knowledge from collaborating actors, thus, our interest is in the socialization and externalization processes.

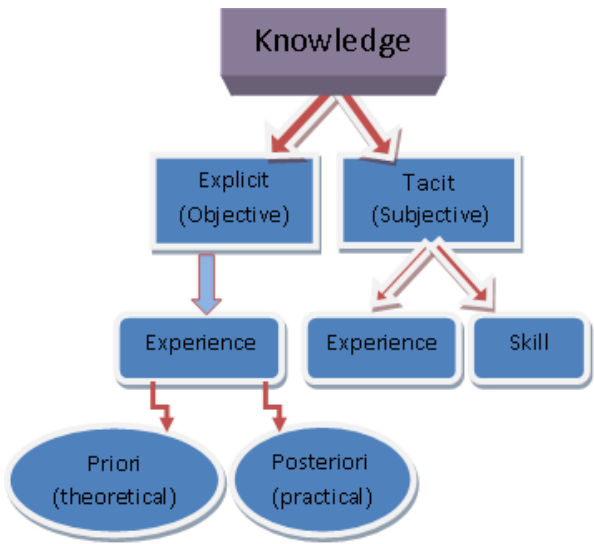

Fig. 1 Classification of Knowledge

TABLE 1
KNOWLEDGE CONVERSION PROCESS

| Process | Instances of Knowledge Transition in EI | Knowledge Conversion |
|---|---|---|
| Socialization | Collaboration for understanding decision problem (DP) | Tacit →Tacit |
| Externalization | Elicitation and definition of DP | Tacit →Explicit |
| Internalization | Analysis of relevant information | Explicit →Tacit |
| Combination | Generation of indicators from relevant information | Explicit →Explicit |

### C. Collaborative Knowledge Management

Knowledge Management (KM) is referred to as a global process in an enterprise which includes all the processes that allow capitalization and sharing of the evolution of knowledge resources (capital) of the firm [7]. Knowledge cannot be dissociated from the context in which it is produced. The individuals or group of individuals (knowledge producers) and their associated socio-cultural environment constitute the context of knowledge production. Hence, KM considers organizational culture in its implementation. It enables people of common vision and culture to share their knowledge, learn from one another and to innovate by working effectively. KM therefore is collaborative in nature.

There are other factors apart from culture which should be inculcated in KM implementation in organizations. These factors are organizational goals, technology and legal protection of Knowledge Resources (*KR*) [12], [13], [14]. Organization can benefit from KM approach if they define their goals and organizational ethics, modus operandi, and personnel's communication forum, including legal measure for protection of knowledge. The technology issue has to do with the implementation of KM system. This requires that a knowledge repository be developed.

*KR* to be capitalized is stored in a repository commonly called "corporate memory (CM) or organizational memory" [14]. CM refers to a structured set of *KR* related to a firm's experience in a given domain. It is essential to identify the goal of KM in order to determine the required kind of CM that would "support the integration of resources and know-how in the enterprise and the co-operation by effective communication and active documentation" [15]. The firm's *KR* could be formalized and modeled in a CM for re-use and update by designated actors. A successful assimilation of technology [16] into KM yields software solution such as, intelligent documentary system, knowledge-base, case-based system, web-based system or a multi-agent system. KM systems enable firms that develop and leverage *KR* to achieve greater success than firms who are more dependent on tangible resources [17]. Some benefits of KM are listed below.

➢ preservation of *KR* from actors and organizational process;
➢ harmonization of firm's explicit and tacit knowledge;
➢ reuse of past project experience or lessons; and
➢ sharing of knowledge to facilitate learning and communication amongst organizational personnel.

Techniques of capitalization of *KR* handle basic required processes such as knowledge acquisition, representation or modeling and exploitation. Knowledge can be elicited and acquired from actors or domain experts through interviews and/or from relevant technical documents. According to [18], KM method involves four stages, namely:

➢ Knowledge Generation that is, knowledge creation and acquisition.
➢ Knowledge Codification that is, internalization or storing of knowledge.
➢ Knowledge Transfer or sharing.

➢ Knowledge Application or reuse.

*D. Review of existing KM systems*

KM is applicable to various disciplines such as industries, medicines, institutions and government. Various KM systems have been developed in these disciplines. We shall therefore consider a few numbers of such applications.

The capitalization of knowledge on how to synthesize "purely Swiss" vitamin C with emphasis on its impact on society through the influence of technology was presented in [19]. In [20], a system on capitalization of design process was discussed. This system focused on the development of design project memory from designers' activities through direct extraction and tracking of knowledge from project design tools, design process and product data. Other applications are capitalization of business experience and resources [21], capitalization of steel production process and defect [22], capitalization of industrial systems [23], capitalization of equipment diagnosis and repair help system [24]. More KM applications shall be reviewed in subsequent sections.

- *Application of Knowledge Management to System Development*

The work focuses on storage and generation of schedules, documents and reports on activities, projects, for instance, Information technology (IT) projects and project management (PM), for sharing of information among project managers [24]. This approach is to prevent reworking and to improve customers' confidence. A web system called 'ProjectWeb' is used as a tool for implementing the KM concept. It is integrated with other specialized tools for progress and quality management in software projects. The methods of exploitation are communication over web mail system and exploration of the library of daily deliverables with PM tools. This work is domain specific.

- *Knowledge formalization in experience feedback processes: An ontology-based approach*

In [25], an Experience Feedback Processes (EFP) model for capitalization of knowledge on changes and improvement of industrial system with the goal of transforming know-how to explicit knowledge on description of learnt lessons is presented. Conceptual graphs are used to model experiences and a formal ontology was defined to describe the concepts and relations that exist among them. Consequently, the domain actors can retrieve for a new problem, similar past experience from industrial experience base with the aid of search reasoning method like projection. The work is domain dependent.

- *Designing a Knowledge Management Approach for the CAMRA Community of Science*

The KM principles namely, technology, human intervention and domain structure were applied to the design of KM system for Center for Advancing Microbial Risk Assessment (CAMRA). CAMRA gathers a community of scientists that investigate several stages in the life cycle of bacterial agents of concern [26]. The knowledge artifacts are referred to as 'Learning Units' (LU), and they are captured in form of research activity, contexts, contributions, and results elements. They proposed the development of repository, web-based KM system with application of technology. This is complemented by knowledge facilitators who mobilize, guide users towards goals and usage of the system; and verify their input for its compliance to the LU. The domain structure handles the integration of users' contribution through the association of new LU with existing ones. The authors claim that this feature ensures knowledge sharing among community members. This approach is recommendable with respect to the emphasis on human factors for successful KM system.

In the reviewed existing systems, handling the interaction among actors is less emphasized. Hence, the approach of KM which facilitates the capitalization of knowledge on actors' communication and its validation process of establishing their concession as well as knowledge on their tasks and solutions is considered in section III.

III. RECURSIVELY DYNAMIC CAPITALIZATION METHOD

We observe that existing KM approach allows acquisition of *KR* as regards a project only after such project might have been completed. We therefore consider such approach as being static in nature. KM should not focus only on explicit knowledge but should also aim at capitalizing actor's tacit knowledge externalized in the course of their interaction. In EI context where actors come together to share, communicate and validate opinions for decision problem resolution, a continuous exchange of knowledge is needed before concession could be reached. The knowledge expressed during the concession process as well the final agreed upon knowledge need to be capitalized. Thus, we propose a dynamic KM approach which enables knowledge in recurrent communication for validation purpose to be capitalized.

The dynamic KM approach as presented in figure 2 adds a new dimension to the conventional KM approach. This dimension addresses the challenge about subjectivity of knowledge in actors' communication during resolution of a problem. There are five major phases in this approach. Each phase is recurrently handled in order to validate *KR* by actors hence leading to evolution of *KR*. Each phase is discussed in turn.

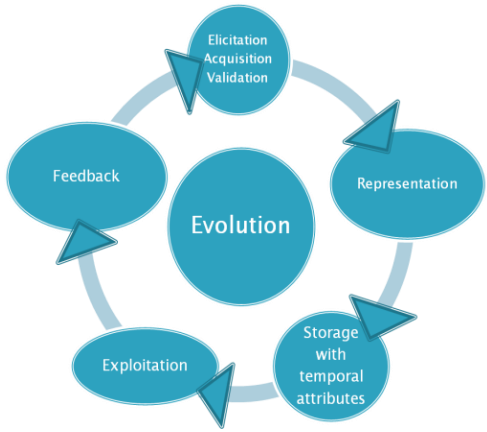

Fig. 2: KM Approach for Socialization and Externalization Processes

1. *Knowledge Elicitation, Acquisition and Validation*: This is the process whereby *KR* are elicited and captured based on the role of individual actors. Thus, we use a case-based algorithm as knowledge acquisition method. In order to capture the tacit knowledge of actors, the communication (socialization process as in table 1) among them is continuously tracked. The externalization of such knowledge requires the process of declaration and annotation. By declaration process, it implies actors have to declare the knowledge with respect to their roles and context at hand. The annotation process captures the recurrent communication among actors for the purpose of proper understanding and validation of *KR*.

The acquired *KR* undergoes validation to ensure high degree of certainty or reliability of such *KR* as conceded by actors. Thus, this phase basically requires a recursive process of declaration and annotation to capture *KR* in such a way that those previously captured are not replaced by new entry. Each entry of *KR* can be distinguished by the respective timestamp. Figure 3 illustrates this phase further using '$T_D$' and '$T_A$' to denote timestamp of *KR* declaration and annotation respectively.

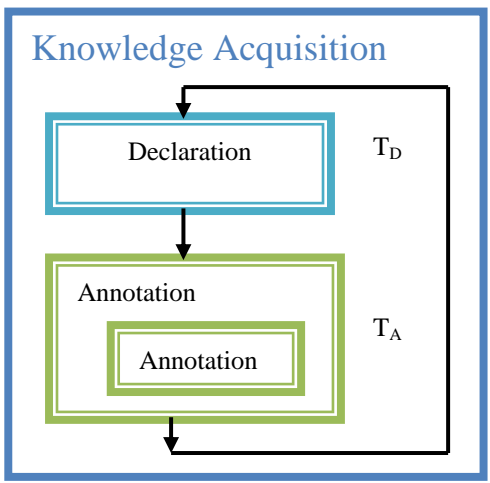

Fig. 3: Recursive Knowledge Acquisition process

**2.** *Knowledge Representation*: *KR* is represented with the aid of a conceptual knowledge model. This model structures *KR* and its properties or description and existing relationship. The generic conceptual model depicted in figure 3 below represents the *KR* required in problem solving process in a given domain. Each component of the model can further be decomposed into more *KR* with regards to the given domain.

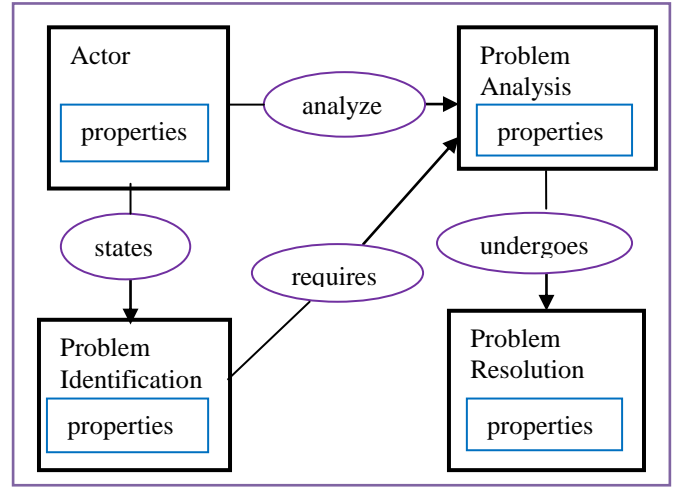

Fig. 3: Generic conceptual Knowledge Model

**3.** *Storage with Temporal Attributes*: The acquired *KR* are stored with temporal attributes that is, date/time stamp in a CM which is called Knowledge Repository in the context of this work. *KR* in current projects are dynamically stored at real time as actors collaborate. The Knowledge Repository is structured such that date/time stamps are stored with each instance of *KR*.

**4.** *Knowledge Exploitation*: The storage of *KR* for instantiation of a Knowledge Repository facilitates knowledge reuse and sharing through exploitation process. We adopt EQUA²te (Explore, Query, Analyze and Annotate) model [27] for knowledge exploitation. EQUA²te tailors the query terms by proposing attributes of the content of Knowledge Repository to aid retrieval of relevant knowledge that meets actors' needs. The search algorithm adopts case-based reasoning approach for retrieval of different cases of knowledge resources such as, participatory actors, process of resolution of decision problem and the result of various activities.

**5.** *Feedback Exploitation Strategy*: This process allows capturing actor's assessment of relevance of exploited knowledge for new problem cases. This strategy enables the evaluation and evolution of the system in such a way that actors can further exploit the generated feedback. This implies that exploitation can lead to feedback generation which subsequently can be exploited.

Thus, the system becomes dynamic as a result of continuous update from knowledge reuse. This KM approach is applied to capitalization of *KR* in the context of DP resolution following EI process.

## IV. IMPLEMENTATION OF THE RECURSIVELY DYNAMIC CAPITALIZATION APPROACH

This section presents the architecture for capitalization of knowledge based on the KM approach presented above. The architecture guides the formalization of the capitalization and exploitation processes. Its implementation yields visualization of knowledge in a collaborative environment.

*A. Architecture for Capitalization of Knowledge*

Figure 4 depicts the architecture for the representation, acquisition, storage and exploitation of *KR* required in DP resolution in a domain. It is based on the proposed KM approach. It presents a platform for representing knowledge of a specific domain (ontology) which can be referenced in relation to *KR* for DP projects following the EI process. It models actor's knowledge in terms of: "*Who*" does "*What*", "*When*", "*Why*", "*How*", and resulting solution or decision. "*When*" is in form of Date/timestamp of respective *KR*. In figure 3, 'T' denotes date/timestamp of *KR* and the subscripts i, i+n, f, for initial, subsequently and final respectively. It serves the purpose of tracking acquisition of distinct *KR* during the collaboration among actors. This feature aids the classification of *KR* according to the validation process and period of cases of projects. The architecture also presents the module for *KR* exploitation. This module is connected with feedback component which allows for acquiring and storing reusable *KR* for new problems. Actors can mine and visualize knowledge from the Knowledge Repository with reference to the related *KR* in the domain ontology. With the temporal attributes, it is possible to exploit *KR* categorized by period of its creation, for example, based on yearly classification.

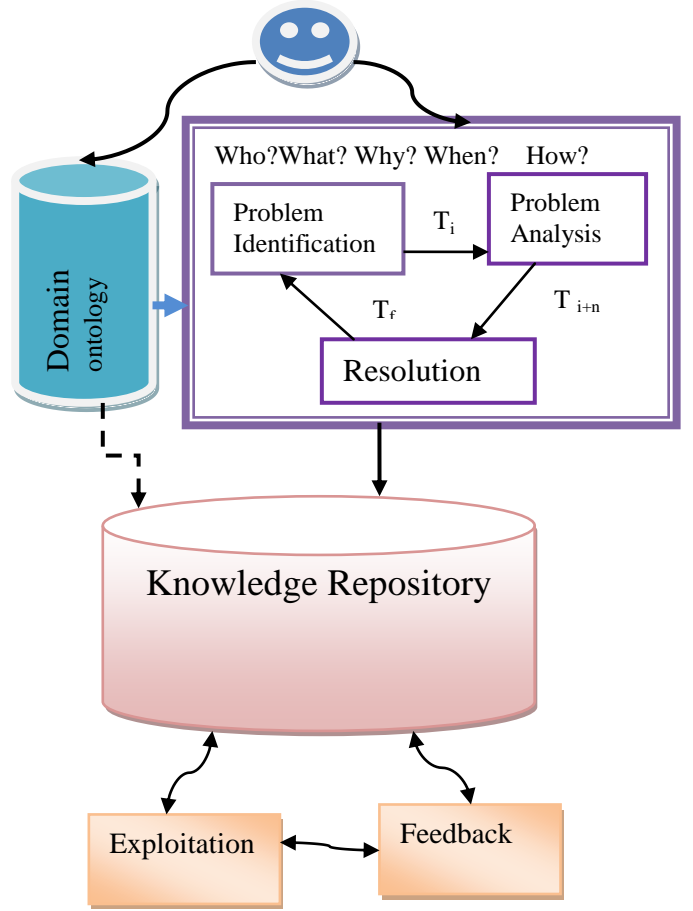

Fig. 4: Architecture for Capitalization of Knowledge

There are two main parts in the architecture. The part above 'Knowledge repository' is for *KR* capitalization while the part below is for exploitation. These two components require a formal description in order to codify a KM system for EI projects.

*B. Formalization of Knowledge Capitalization Method*

The capitalization of *KR* is defined in form of a function *F* denoted as:

$$F(v, w, x, y, z, t) \Rightarrow \lambda(R) \qquad (1)$$

Where:
v ϵ V: who
w ϵ W: what
x ϵ X: why
y ϵ Y: how
z ϵ Z: result/solution
t ϵ T: when (date/timestamp)

and V, W, X, Y, Z ,T : sets of knowledge resources(*KR*)
R: set of aggregated *KR*
$\forall$ v, w, x, y, z, t $\exists$ $F(v, w, x, y, z, t) \Rightarrow \lambda(R)$ (2)
Where:

λ : capitalization process on sets of (v, w, x, y, z, t) to obtain R

Capitalization process: functional methods of knowledge aggregation operations

Aggregation operations:- {|Acquisition|, |Annotation|, |Acquisition|} //*KR* are acquired before and after annotation

[$F_1$, $F_2$..$F_n$] are sets of aggregated *KR* that are captured based on actors' roles using case based algorithm.

In (2) above, collection of v, w, x, y, z, and t for a specific case of decision problem project implies there is a function *F* that is evaluated as or equivalent to aggregated *KR* denoted by R.

The exploitation process is subsequently analyzed.

*C. Formalization of Knowledge exploitation method*

The exploitation part of the architecture depicts retrieval of knowledge for visualization in a way that reduces irrelevant search results with respect to specific actor's needs. This is realized with the use of query terms that are representative of the attributes of the content of the Knowledge Repository. Query processing is handled by the search algorithm or query-result matching algorithm presented below.

Given,
$$F(v, w, x, y, t, z) \Rightarrow \lambda(R) \text{ as in (1) above,}$$
Let $\lambda(R) \Rightarrow \lambda(S)$
Where:
S = {set of aggregated *KR*}
∃ s: s ϵ S
Let Qt = {set of the query terms/attributes}
And a, b, c, be elements of Qt
a, b, c, ϵ Qt
We define a function δ, where δ denotes the combination of elements of Qt
Such that,
$$\delta(a, b, c) \Leftrightarrow s \epsilon R \quad (3)$$
$$\text{Thus, } \lambda(s) \neg \text{ null} \quad (4)$$

In (3), for every combination of query terms, there is a search result- s retrieved from the capitalized *KR* in R. The implication of this is that exploitation process returns values from the set of aggregated *KR* as concluded in (4).

In order to validate the KM approach, 'Recursively Dynamic Algorithm' for implementing a KM System in EI is presented subsequently.

*D. Recursively Dynamic Algorithm of Capitalization of Knowledge*

Technology aids in the codification of *KR* capitalization. Thus, this work proposes 'Recursively Dynamic Algorithm' based on the proposed methods for the implementation of a KM system which allows dynamic capitalization and visualization of knowledge by actors. The design tool for specifying the algorithm is Pseudo-codes. This prepares the codification of the KM system and serves as a plan for the system implementation. It is documented as capitalization and exploitation processes respectively.

**Capitalization-process** ()
**Foreach** EI project E
1. Select CASE (actor-role) to specify current EI task
Repeat
Call annotation function ()
2. Specify/validate knowledge on EI task
3. Track the timestamp
4. Store 1, 2 and 3 into knowledge repository
Until concession given by actor(s)
**Endfor**

The pseudo-code above describes the capitalization process for knowledge acquisition from actors. λ in (1) corresponds to this process such that λ(R) implies the resultant set of related *KR* in a Knowledge Repository. The process uses Case based reasoning method to determine what function to invoke according to actor's role and the task to be performed. With reference to an EI project, if a DP is to be initiated by an actor having as role - decision maker, the system will respond by invoking the function for DP externalization into document.

The function 'annotation' captures the communication between the actors at each specific phase of resolution of a DP by recursively invoking the annotation function for each actor. The watcher proposes his knowledge on the problem resolution while the decision maker evaluates and validates his (watcher) proposal in the context of the DP at hand. The process is continuously executed until the decision maker finally approves or validates such knowledge. Thus, the dynamic feature is realized by the iterative tracking and storage of *KR* before concession is reached. Subsequently, there is need for exploitation process which will enable actors to exploit *KR* for solving related new problems.

**Exploitation process** ()
Exploitation-method == {|explore| |query|…}
While not End(exploitation)
  IF (Exploitation-method == explore)
    Then
            Browse cases of classified *KR* from
            knowledge repository
  Else
      Begin
        Specify query terms
         Execute combination process
          Match query terms with knowledge repository
          content
         Visualize *KR* {|validated|
          |evolution| |complete|}
      End
  End (IF)
End (While)

The exploitation process describes the method of exploiting and visualizing capitalized knowledge. It is based on the EQUA$^2$TE (Explore, Query, Analyze and Annotate) method and case based reasoning matching technique. It is of two modes, namely, exploration and querying modes.

Function δ in (2) is executed by this algorithm in a way that facilitates visualization of *KR* by actors through query formulation. The other option is through direct exploration of clusters of related *KR*. Visualization of *KR* is classified into three options. They are - validated *KR*, its evolution before the validation and the complete sequence from initiation through to validation. The next section describes the implementation of the algorithm.

*E. KM System Testing and Application to Case Study*

The Recursively Dynamic Algorithm for Capitalization of Knowledge was implemented in the form of a web-based prototype called EIKC (Economic Intelligence Knowledge Capitalization) system. The sample case study was Sunseed Oil Nigeria Plc. (Private Limited Company). The survey of the case study was carried out as a Professional Master Degree project. The board of directors of the company constitutes the decision makers while the product researchers serve as watchers. They normally use the general managerial approach for handling DP. However, the application of EI process was proposed to them. A typical DP on how to improve the productivity of the company and to guarantee customer satisfaction was considered in the light of total automation of production and sales processes.

EIKC system is simulated with a number of scenarios of DP projects including that of total automation specified for our case study. Knowledge is capitalized from actors according to the EI process in the course of resolving the DP. The snapshots in figures 5 and 6 describe the knowledge capitalization of DP resolution on 'total automation of production and sales processes'.

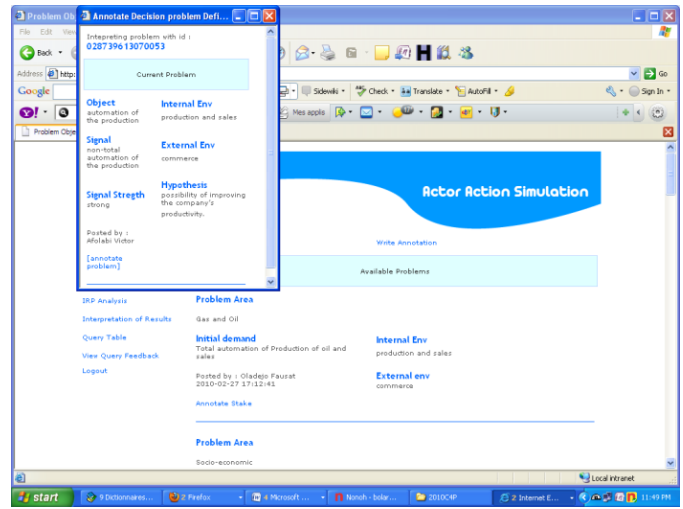

Fig. 6: Validation of Decision Problem definition

In figure 7, the system was accessed by the decision maker to present DP and its contexts in form of internal and external environments. This DP document was automatically stored with the timestamp. Subsequently, the watcher defines the stake of the DP as illustrated in figure 8. Decision maker used annotation interface for the validation of the specified stake while the timestamp is also stored. The watcher reacts to decision maker's annotation and update DP stake accordingly. This process is repeated until concession is reached. The knowledge of the subsequent tasks of the resolution process was capitalized with respective links on the EIKC system interface.

Actors exploit the KM system by visualizing both validated and pre-validated or evolving knowledge of specific tasks of resolving DP according to EI process. Figure 7illustrates knowledge of actors on the DP as it evolved. It also makes visible the actors who originated such knowledge and the respective timestamps.

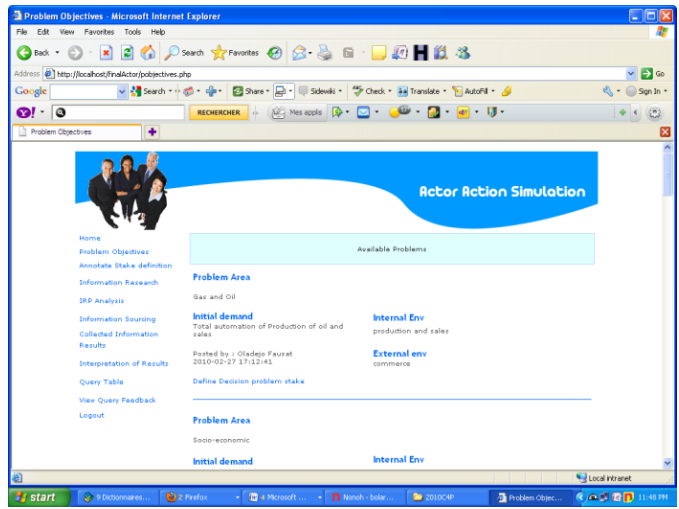

Fig. 5: Capitalization of Decision Problem declaration

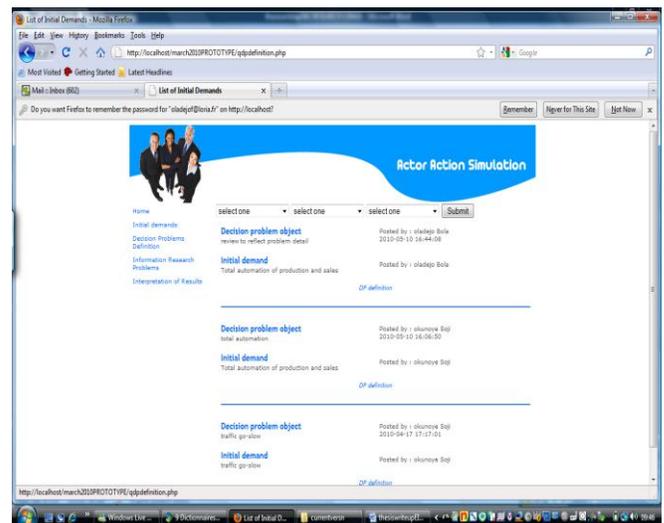

Fig. 7: Visualization of knowledge

The implication of the proposed KM approach and EIKC system is that creation and exploitation of collaborative knowledge of actors in the course of handling DP resolution yield validated or reliable knowledge which is accessible to them based on their needs.

V. CONCLUSION

In this paper, we explained a recursively dynamic capitalization or KM approach for managing actors' knowledge expressed in decision problem solving activities. EI process was described as a framework for strategic resolution of decision problem.

Recursively dynamic algorithm for capitalization of knowledge was implemented in the form of a web-based prototype called EIKC system. We conclude that the approach of dynamic capitalization of knowledge supports knowledge validation process among collaborating actors, hence increasing the reliability of capitalized knowledge. Visualization of knowledge is presently in form of navigation among categories of related *KR* or knowledge taxonomy. We are currently working on optimizing the visualization aspect of EIKC system to cater for graphical display of knowledge network.